\documentclass{article}

   \usepackage[nonatbib, final]{neurips_2021}

\usepackage[utf8]{inputenc} 
\usepackage[T1]{fontenc}    
\usepackage{hyperref}       
\usepackage{url}            
\usepackage{booktabs}       
\usepackage{amsfonts}       
\usepackage{nicefrac}       
\usepackage{microtype}      
\usepackage{xcolor}         
\usepackage{graphicx}
\usepackage{bm}
\usepackage{physics}
\usepackage{multirow}
\usepackage[square, comma, sort&compress, numbers]{natbib}

\title{Exploring Student Representation For Neural Cognitive Diagnosis}

\author{%
  Hengyao Bao \\
  Tencent Inc.\\
  Chengdu, China\\
  \texttt{henniebao@tencent.com} 
  \And
  Xihua Li \\
  Tencent Inc.\\
  Chengdu, China\\
  \texttt{lixihua9@126.com} 
  \And
  Xuemin Zhao \\
  Tencent Inc.\\
  Chengdu, China\\
  \texttt{xueminzhao@tencent.com} 
  \And
  Yunbo Cao \\
  Tencent Inc.\\
  Beijing, China\\
  \texttt{yunbocao@tencent.com} 
}

\begin{document}

\maketitle

\begin{abstract}

Cognitive diagnosis, the goal of which is to obtain the proficiency level of students on specific knowledge concepts, is an fundamental task in smart educational systems. 
Previous works usually represent each student as a trainable knowledge proficiency vector, which cannot capture the relations of concepts and the basic profile(e.g. memory or comprehension) of students.
In this paper, we propose a method of student representation with the exploration of the hierarchical relations of knowledge concepts and student embedding.
Specifically, since the proficiency on parent knowledge concepts reflects the correlation between knowledge concepts, we get the first knowledge proficiency with a parent-child concepts projection layer.
In addition, a low-dimension dense vector is adopted as the embedding of each student, and obtain the second knowledge proficiency with a full connection layer. Then, we combine the two proficiency vector above to get the final representation of students. Experiments show the effectiveness of proposed representation method.

\end{abstract}

\section{Introduction}

\begin{figure}[htb] \label{cd-example}
  \centering
  \includegraphics[width=0.85\textwidth]{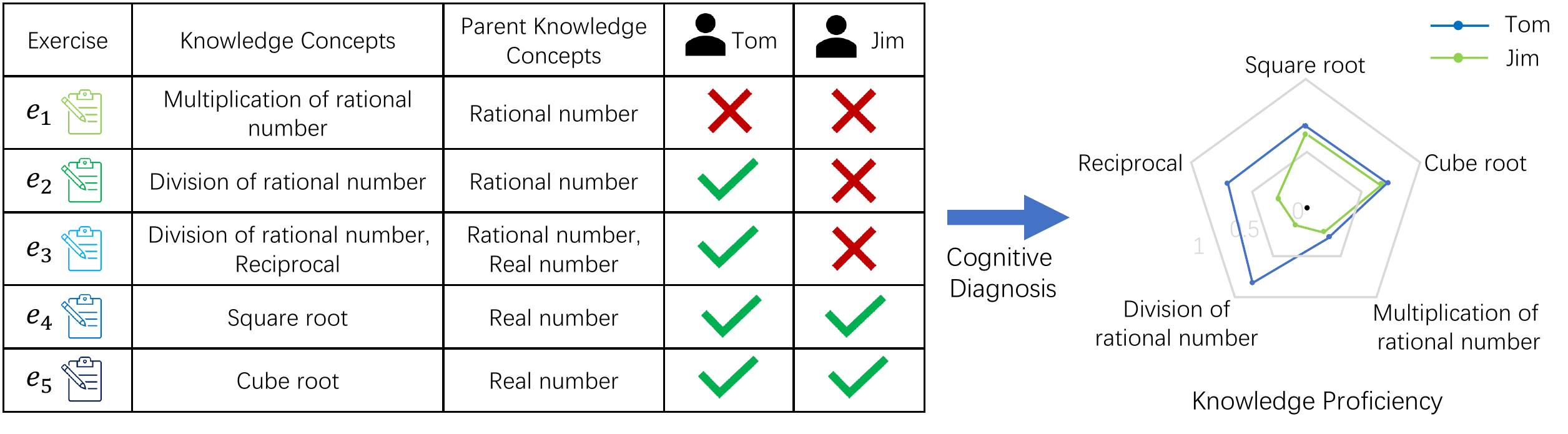}
  \caption{An example of cognitive diagnosis}
\end{figure}

Cognitive diagnosis is an essential and fundamental technology in smart education systems, in which cognitive diagnosis can help to obtain the proper profiles of students and assist lots of education services, such as student learning report and adaptive exercise recommendation \cite{kuh2011, lana}.
Figure 1 shows an example of cognitive diagnosis process.
Generally, with the exercise responses of students and labels of exercises, cognitive diagnosis is to infer their relative abilities \cite{irt}, such as proficiency on specific knowledge concepts (e.g. \emph {multiplication of rational number}) \cite{ncd}.

Many classical methods have been developed to address this issue, such as Multidimensional Item Response Theory (MIRT) \cite{mirt}, Deterministic Inputs, Noisy And gate model (DINA) \cite{dina}, Matrix Factorization (MF) \cite{mf}, and Item response ranking framework \cite{irr}.
Recently, deep neural network has also been applied in cognitive diagnosis.
\cite{ncd} proposed a Neural Cognitive Diagnosis framework (NCD) which utilizes a knowledge proficiency vector to represent student and formulates the students, exercises and responses with an MIRT-like multi-layer perceptron.
In \cite{ecd} an Educational context-aware Cognitive Diagnosis framework (ECD) was developed to model the context of student, e.g. highest education degree of parents and duration in early childhood education.
However, since only a knowledge proficiency vector is used to represent the student, these methods are short of characterizing the complete profile of student, such as the comprehensive ability of student, or the average mastery on associated knowledge concepts.
For Example, as shown in Figure 1, although Tom and Jim both answer $e_1$ incorrectly, it's still not suitable for the cognitive diagnosis system to give similar poor scores on concept \emph{multiplication of rational number}, since Tom is obviously a good student and has a better performance in the domain of \emph{rational number}.

In this paper, we develop a structure to enrich the representation of students by making use of the hierarchical relations of knowledge concepts and the embedding of students. 
First, the proficiency on parent knowledge concepts is used to represent the knowledge related profile of students, and the proficiency on child concepts is obtained with a parent-child concepts projection layer.
Second, for a further exploration on the representation of students, 
we adopt a low-dimension dense vector as the embedding of each student, and obtain the second knowledge proficiency with a full connection layer.
Then, we average the two proficiency above to get the final representation of students, and formulate the students, exercises and responses as a neural diagnosis network. Experiments show that it has a substantial improvement in terms of both response prediction and knowledge proficiency diagnosis.

\section{Model}

\begin{figure}[htb] \label{fig_struct}
  \centering
  \includegraphics[width=0.85\textwidth]{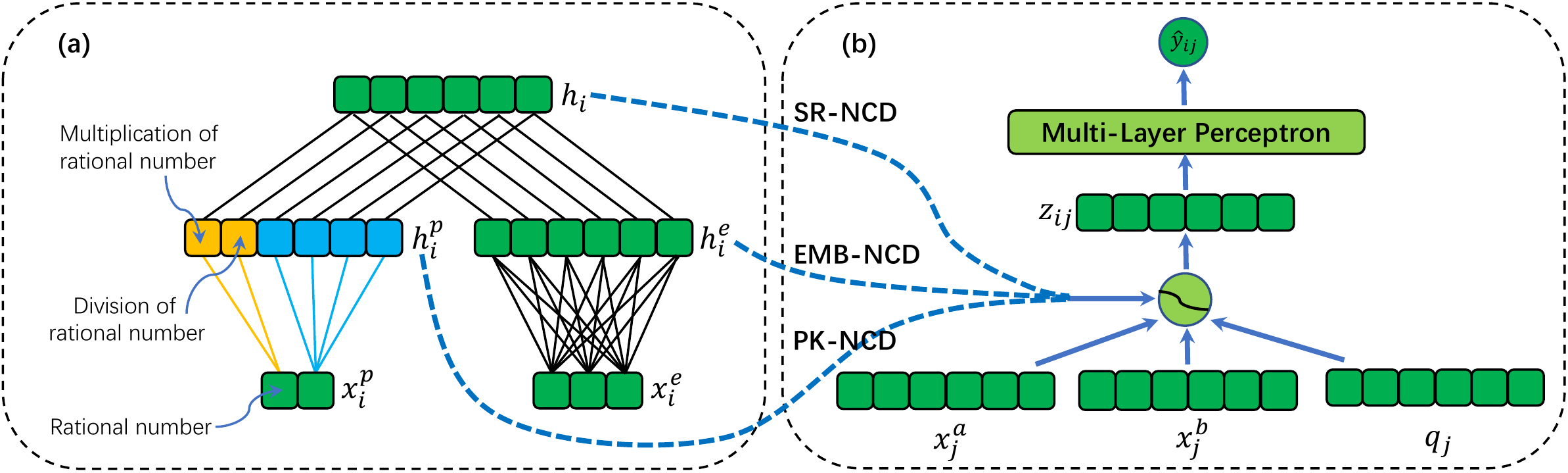}
  \caption{The structure of proposed method: (a) student representation layer, (b) answer correctness prediction layer}
\end{figure}


\subsection{Problem Definition}

Suppose in a smart education system there are $N$ students and $M$ exercises, and define the responses of students as $R = \{\langle s_i, e_j, y_{ij} \rangle | 1 \leq i \leq N, 1 \leq j \leq M, y_{ij} \in \{0, 1\}\}$, where $s_i$, $e_j$ and $y_{ij}$ denote the $i$-th student, $j$-th exercise and the relative response of student i on exercise j respectively. 
In addition, we define the Q-matrix (usually labelled by experts) as $\bm{Q}=\{q_{kj}\}_{K \times M}$, in which $q_{kj} \in \{0, 1\}$ denote whether exercise $e_j$ relates to the $k$-th knowledge concept, and $K$ is the number of concepts. Then, given the responses of students $R$ and the Q-matrix $\bm{Q}$, the goal of cognitive diagnosis is to estimate the knowledge proficiency of each student.

\subsection{Student Representation}

The proposed method of student representation is illustrated in Figure \ref{fig_struct}(a).
We first notice that, the knowledge concepts to be diagnosed have related parent knowledge concepts that are labeled by experts in advance.
As the case shown in Figure \ref{cd-example}, \emph {rational number} is the parent knowledge concept of both \emph {multiplication of rational number} and \emph {division of rational number}.
Generally, the proficiency in parent knowledge concepts can somehow indicate students' knowledge related profile and mastery in child knowledge concepts.\footnote{Without specification, the term \emph{knowledge concepts} or \emph{child knowledge concepts} denotes the leaf nodes of concept tree, and \emph{parent knowledge concepts} denotes the parent nodes of leaf nodes.}
Therefore, the parent-child relations of knowledge concepts can be used to enrich the representation of students.
Suppose the $K$ knowledge concepts has $L$ parent concepts, and we use a trainable vector $\bm{x}^p_i \in \mathbb{R}^{L\times1}$ to represent the proficiency of $i$-th student in each parent concepts. Then the knowledge proficiency in child concepts $\bm{h}_i^p$ can be obtained by: 
\begin{align}
\bm{h}_i^p = \sigma (\bm{G}\bm{x}_e^p + \bm{b}^p) \label{hpi}
\end{align}
where $\sigma(\dotproduct)$ denotes the sigmoid function, $\bm{b}^p \in \mathbb{R}^{K\times1}$ is a bias vector, and $\bm{G}=\{g_{kl}\}_{K\times L}$ is a parent-child map matrix in which $g_{kl}$ is a trainable variable if the $k$-th child concept is descendent of the $l$-th parent concept, and $g_{kl}\equiv0$ otherwise.

In the other hand, $\bm{h}_i^p$ only utilizes the relations of same knowledge concept family, and cannot fit the proficiency of different concept families,  or the memory and comprehensive ability of students. Hence, we use a low-dimension dense vector $\bm{x}^e_i \in \mathbb{R}^{D\times1}$ as the embedding of student $s_i$, and get the knowledge proficiency $\bm{h}_i^e$ by:
\begin{align}
\bm{h}_i^e = \sigma (\bm{F}\bm{x}_i^e + \bm{b}^e) \label{hei}
\end{align}
where $\bm{b}^e \in \mathbb{R}^{K\times1}$ is a bias vector, and $\bm{F} \in \mathbb{R}^{K\times D}$ is a projection matrix to knowledge concepts.

Then, we can get the final knowledge proficiency $\bm{h}_i$ by simply calculating the mean:
\begin{align}
\bm{h}_i = (\bm{h}_i^p + \bm{h}_i^e )\ /\ 2 \label{hi}
\end{align}
Note that, like the work in \cite{ecd}, one also get $\bm{h}_i$ by a weighted sum of $\bm{h}_i^p$ and $\bm{h}_i^e$, in which the weight is also trainable. However, we do not see it has a substantial improve in our dataset.

\subsection{Answer Correctness Prediction} \label{section_pred}

With the knowledge proficiency obtained above, the task of cognitive diagnosis can be formulated as an answer correctness prediction problem \cite{ncd}.
The structure of prediction layer is shown in Figure \ref{fig_struct}(b).
Specifically, for the exercise $e_j$, we define $\bm{q}_j $ as the $j$-th column of Q-matrix $\bm{Q}$, $\bm{x}^a_j \in \mathbb{R}^{K\times1}$ and $\bm{x}^b_j \in \mathbb{R}^{K\times1}$ as the discrimination and difficulty embedding respectively, and the prediction of answer correctness $y_{ij}$ is obtained by:
\begin{align}
& \bm{\alpha}_j = \sigma (\bm{x}^a_j) \\
& \bm{\beta}_j = \sigma (\bm{x}^b_j) \\
& \bm{z}_{ij} = \bm{q}_j \dotproduct \bm{\alpha}_j \dotproduct (\bm{\beta}_j - \bm{h}_i) \label{zij} \\
& \hat{y}_{ij} = MLP(\bm{z}_{ij}) \label{mlp} 
\end{align}
where $MLP(\dotproduct)$ denotes a multi-layer perceptron, and $\hat{y}_{ij}$ is the prediction result. 
Thus the cross entropy loss for student $s_i$ on exercise $e_j$ is defined as:
\begin{align}
loss_{ij} = y_{ij}\text{log}\,\hat{y}_{ij} + (1-y_{ij})\text{log}\,(1-\hat{y}_{ij})
\end{align}
In addition, to satisfy the \emph{monotonicity assumption} \cite{mirt} to ensure good performance and interpretability, we {}restrict $\bm{G}$ in \eqref{hpi} and weight matrix of the multi-layer perceptron \eqref{mlp} to be positive when training \cite{ncd}. Thus, the higher each entry of $\bm{h}_i$ or $\bm{x}_i^p$ is, the more likely the student answers the exercise correctly.
Also note that both the knowledge proficiency $\bm{h}^p_i$ and $\bm{h}^e_i$ can also be passed to \eqref{zij} independently for prediction (for reason of same dimentions and same representation abilities). 
As shown in Figure \ref{fig_struct}(b), for simplification, we denotes the method using the parent knowledge $\bm{h}^p_i$ as PK-NCD($\bm{P}$arent $\bm{K}$nowledge), method using the student embedding $\bm{h}^e_i$ as EMB-NCD($\bm{EMB}$edding), and method using the student representation $\bm{h}_i$ as SR-NCD($\bm{S}$tudent $\bm{R}$epresentation) respectively.

\section{Experiments}


\subsection{Datasets, Metrics and Setups}

We test the cognitive diagnosis models with two datasets of real-world education scenarios, i.e. ASSIST \cite{assist} and XCLASS-MATH. See the datasets details in Appendix \ref{datasets}. Besides, since there are no ground-truth values for the knowledge proficiency of students, it is difficult to evaluate the models straightforwardly. Following the work in \cite{irr}, we evaluate the performance of models from two perspectives. First, we use Accuracy (ACC) and Area Under the Curve (AUC) to test the classification abilities of models. Second, we adopt Degree Of Agreement (DOA) to assess the monotonicity of models. See the definition of DOA in Appendix \ref{doa}.

We evaluate the proposed EMB-NCD, PK-NCD and SR-NCD defined in Section \ref{section_pred} in the experiments.
Since ASSIST does not have information about parent knowledge concepts, only EMB-NCD is tested in its experiments.
Beside, we adopt two hidden layers in the MLP \eqref{mlp}, and set the dimension as 512, 256 for ASSIST, and 128, 64 for XCLASS-MATH respectively, and the dimension of student embedding $D$ is set to $K/4$.
We also compare the performance of proposed methods with several previous works: DINA\cite{dina}, MIRT\cite{mirt}, NCD\cite{ncd}.

\subsection{Resutls}

\begin{table}[htb]
  \caption{Experimental results}
  \label{results}
  \centering
  \begin{tabular}{lllllll}
    \toprule
    \multirow{2}*{Model\quad\quad} 
    &\multicolumn{3}{c}{ASSIST} &\multicolumn{3}{c}{XCLASS-MATH}  \\
    \cmidrule(r){2-4} \cmidrule(r){5-7}
     & ACC     & AUC  &DOA & ACC & AUC  &DOA \\
    \midrule
    DINA & 0.682  & 0.727 & 0.603  & 0.670  & 0.712 & 0.629 \\
    MIRT & 0.724  & 0.733 & 0.601  & 0.746  & 0.754 & 0.632 \\
    NCD & 0.726  & 0.757 & 0.609  & 0.745  & 0.763 & 0.635 \\
    EMB-NCD & \textbf{0.735}  & \textbf{0.771} & \textbf{0.681}  & 0.748  & 0.768 & 0.658 \\
    PK-NCD & -  & - & -  & 0.753  & 0.768 & 0.656 \\
    SR-NCD & -  & - & -  & \textbf{0.757}  & \textbf{0.780} & \textbf{0.664} \\
    \bottomrule
  \end{tabular}
\end{table}

\begin{figure}[htb] 
  \centering 
  \includegraphics[width=1\textwidth]{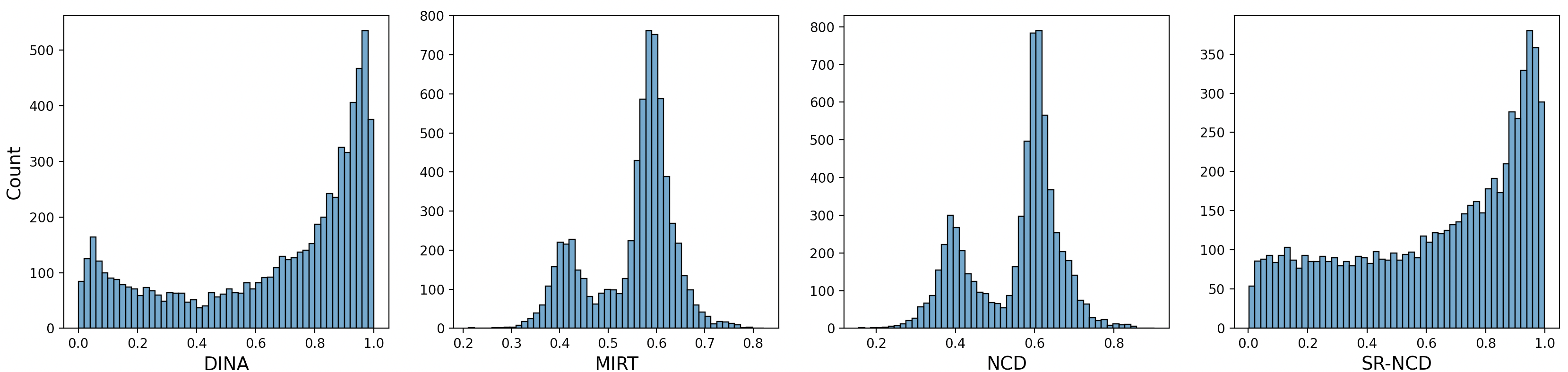}
  \caption{Distribution histogram of knowledge proficiency}
\end{figure}

The experimental results are shown in Table \ref{results}. One can observe that the proposed methods outperform all the other baselines on both datasets. Specifically, even though simply adding a student embedding layer, EMB-NCD can obtain a significant improvement compared with the original NCD. Furthermore, PK-NCD and EMB-NCD have similar performance, and by combining them we can acquire another obvious gain. 
Thus, the experimental results demonstrate the effectiveness of the proposed student representation methods.

Meanwhile, we also display the distribution histogram of knowledge proficiency obtained from XCLASS-MATH in Figure 3. 
It's interesting to notice that the knowledge proficiency of MIRT and NCD have almost the same distribution, since MIRT is a special case of NCD \cite{ncd}. 
Besides, the distribution curve of DINA, MIRT and NCD are bimodal. On the contrary, SR-NCD has a convex and much more smooth curve, which indicates the proficiency acquired might be more discriminative.

\section{Conclusion}

In this paper, we considered the problem of student representation in cognitive diagnosis model.
We developed a method of student representation with the exploration of the hierarchical relations of knowledge concepts and student embedding.
Experiments demonstrate the effectiveness and interpretability of the proposed methods.





\appendix

\section{Datasets} \label{datasets}

\begin{table} [htb]
  \caption{Dataset summary}
  \label{dataset-table}
  \centering
  \begin{tabular}{lll}
    \toprule
    \cmidrule(r){1-3}
    Statistics     & ASSIST\quad\quad      & XCLASS-MATH \\
    \midrule
    \# students                  & 4,163    & 165  \\
    \# exercises                 & 17,746   & 250  \\
    \# knowledge concepts        & 123      & 74  \\
    \# parent knowledge concepts & \ \ /        & 7  \\
    \# response logs             & 324,572  & 13,574  \\
    \# average logs per student  & 77.97    & 82.27  \\
    \# average logs per exercise & 18.29    & 54.30  \\
    \bottomrule
  \end{tabular}
\end{table}
The statistics of the datasets are summarized in Table 2.
ASSIST (ASSISTments 2009-2010 ``skill builder'') is a widely used open dataset collected by the ASSISTments online tutoring systems\footnote{https://sites.google.com/site/assistmentsdata/home/assistment-2009-2010-data/skill-builder-data-2009-2010}.
XCLASS-MATH is a mathematical dataset collected by the smart education system XCLASS\footnote{https://xclass.qq.com. XCLASS-MATH will be released later.}, in which teachers assign and correct students' homework online, and students do their homework with an e-ink pad.
It mainly contains the mathematical homework logs within two months of the $7$-th grade students of an middle school.

\section{Degree of Agreement} \label{doa}

The Degree of Agreement (DOA) is defined as:
\begin{align}
DOA(k)=\frac{1}{Z} \sum_{i=1}^{N} \sum_{j=1}^{N} \delta(h_{ik},h_{jk})\frac{\sum_{l=1}^{M}I_{lk}\land J(l,i,j)\land \delta(y_{il}, y_{jl})}{\sum_{l=1}^{M}I_{lk}\land J(l,i,j)\land [y_{il}\neq y_{jl}]}
\end{align}
where $Z=\sum_{i=1}^{N} \sum_{j=1}^{N} \delta(h_{ik},h_{jk})$, $h_{ik}$ denotes the proficiency of student $i$ on concept $k$, $\delta(x,y)=1$ if $x>y$ and $\delta(x,y)=0$ otherwise, $I_{lk}=1$ if exercise $l$ contains concept $k$ and $I_{lk}=0$ otherwise, and $J(l,i,j)=1$ if both student $i$ and $j$ did exercise $l$ and  $J(l,i,j)=0$ otherwise. The perspective of DOA is that, 
if student $i$ has a higher proficiency on concept $k$ than student $j$, then student $i$ is more likely to answer exercise related to concept $k$ correctly than student $j$. The average of $DOA(k)$ on all concepts is used in our experiments. Thus, the model with a higher DOA score might have a better monotonicity and interpretability.


\clearpage

\begin{thebibliography}{1}
\small

\bibitem{kuh2011} Kuh  G. D., Kinzie J., Buckley J. A., Bridges B. K., and Hayek J. C. (2011) {\it Piecing together the student success puzzle: research, propositions, and recommendations: ASHE Higher Education Report}, volume 116. John Wiley \& Sons.

\bibitem{lana} Yuhao Zhou, Xihua Li, Yunbo Cao, Xuemin Zhao, Qing Ye, and Jiancheng Lv (2021) LANA: towards personalized deep knowledge tracing through distinguishable interactive sequences. In {\it Proceedings of the Educational Data Mining}.

\bibitem{irt} Embretson S. E., and Reise S. P. (2013) {\it Item response theory}. Psychology Press.

\bibitem{mirt} Reckase, M. D.\ (2009) Multidimensional item response theory models. In {\it Multidimensional Item Response Theory}, pp.\ 79-112. Springer.

\bibitem{dina} De La Torre, J. (2009) Dina model and parameter estimation: A didactic. {\it Journal of educational and behavioral statistics} 34(1):115–130.

\bibitem{mf} Koren Y., Bell R., and Volinsky C. (2009) Matrix factorization techniques for recommender systems. {\it Computer} 42(8), pp.\ 30–37.

\bibitem{irr} Shiwei Tong, Qi Liu, Runlong Yu, Wei Huang, Zhenya Huang, Zachary A. Pardos, and Weijie Jiang (2021) Item response ranking for cognitive diagnosis. In {\it Proceedings of the Thirtieth International Joint Conference on Artificial Intelligence}, pp.\ 1750-1756.

\bibitem{ncd} Fei Wang, Qi Liu, Enhong Chen, Zhenya Huang, Yuying Chen, Yu Yin, Zai Huang, and Shijin Wang (2020) Neural cognitive diagnosis for intelligent education systems. In {\it Proceedings of the AAAI Conference on Artificial Intelligence}, volume 34, pp.\ 6153–6161.

\bibitem{ecd} Yuqiang Zhou, Qi Liu, Jinze Wu, Fei Wang, Zhenya Huang (2021) Modeling context-aware features for cognitive diagnosis in student learning. In {\it Proceedings of the 27th ACM SIGKDD Conference on Knowledge Discovery \& Data Mining}, pp.\ 2420–2428.

\bibitem{assist} Feng M., Heffernan N., and Koedinger K. (2009) Addressing the assessment challenge with an online system that tutors as it assesses. {\it User Modeling and User-Adapted Interaction} 19(3): 243-266.






\end{thebibliography}
\end{document}